\documentclass{singlecol-new}

\usepackage{natbib,stfloats}
\usepackage{mathrsfs}

\theoremstyle{TH}{

}

\theoremstyle{THrm}{

}

\theoremstyle{THhit}{

}

\makeatletter
\def\theequation{\arabic{equation}}

\usepackage{graphicx}

\usepackage{pgfplots}
\usepackage{tikz}
\usetikzlibrary{arrows,decorations.pathmorphing,fit,positioning}
\usepackage{xcolor}
\usetikzlibrary{patterns}
\usepackage{amstext}

\usepackage{float}

\usepackage{kbordermatrix}

\usepackage[hyphenbreaks]{breakurl}
\usepackage[hyphens]{url}

\usepackage{multirow}

\usepackage{tikz}
\usetikzlibrary{trees}

\tikzset{level 1/.style={level distance=2cm, sibling distance=10cm}}
\tikzset{level 2/.style={level distance=2cm, sibling distance=3cm}}

\tikzset{bag/.style={text width=15em, text centered,yshift=-0.5cm}}

\usetikzlibrary{arrows,decorations.pathmorphing,fit,positioning}

\pgfplotsset{compat=1.5}

\def\addlegendimage{\csname pgfplots@addlegendimage\endcsname}

\makeatother

\begin{document}%

\setcounter{page}{1}

\LRH{A. Karami}

\RRH{Application of Fuzzy Clustering for Text Data Dimensionality Reduction}

\VOL{6}

\ISSUE{3, 2019}

\PUBYEAR{xxx}

\BottomCatch

\CLline

\subtitle{}

\title{Application of Fuzzy Clustering for Text Data Dimensionality Reduction}

\authorA{Amir Karami}
\affA{College of Information and Communications\\ University of South Carolina \\ E-mail: karami@sc.edu}
%
%

%

%
%
%
%
%
%
%

\begin{abstract}
Large textual corpora are often represented by the document-term frequency matrix whose elements are the frequency of terms; however, this matrix has two problems: sparsity and high dimensionality. Four dimension reduction strategies are used to address these problems. Of the four strategies, unsupervised feature transformation (UFT) is a popular and efficient strategy to map the terms to a new basis in the document-term frequency matrix. Although several UFT-based methods have been developed, fuzzy clustering has not been considered for dimensionality reduction. This research explores fuzzy clustering as a new UFT-based approach to create a lower-dimensional representation of documents. Performance of fuzzy clustering with and without using global term weighting methods is shown to exceed principal component analysis and singular value decomposition. This study also explores the effect of applying different fuzzifier values on fuzzy clustering for dimensionality reduction purpose.

\end{abstract}

\KEYWORD{Dimension Reduction; Fuzzy Clustering; SVD; PCA; Term Weighting; Fuzzifier; Classification.}

\REF{to this paper should be made as follows: Karami, A. (2019)
``Application of fuzzy clustering for text data dimensionality reduction",
Int. J. Knowledge Engineering and Data Mining, Vol. 6, No. 3, pp.289-306.}

\begin{bio}
Amir Karami is Assistant Professor in the College of
Information and Communications and Faculty Associate in the Arnold School
of Public Health at the University of South Carolina. His research interests
are text mining, social media analysis, computational social science and
medical/health informatics.\vs{9}

\end{bio}

\maketitle

\section{Introduction}

Very large{-}scale text data management and analysis has been one of the frontiers for current and future research  \citep{council2016future} due to the availability of millions of documents in digital libraries, such as MEDLINE/PubMed \citep{karami2019exploring}, and billions of comments and posts in social media such as Facebook and Twitter discussing different issues such as health \citep{karami2019exploratory,karami2018characterizingtrans}, organizations \citep{karami2018us}, politics \citep{karami2019political}, and natural disasters \citep{karami2019twitter}. While these large datasets provide extremely useful and valuable resources for researchers, there is a need to develop new approaches for processing high dimensional data.

Document-term matrix (DTM) is a common method to represent documents using frequencies of words in a corpus \citep{patel2018fuzzy}. For example, in the following DTM word 3 ($w_3$) appeared 2 times in document 1 ($d_1$).

	\[
	A= \kbordermatrix{
		& w_1 & w_2 & w_3 & w_4 & w_5 & w_6 & w_7 & w_8 & w_9 & w_{10} \\
		d_1 & 1 & 0 & 2 & 0  & 1& 0& 0& 0& 0& 2\\
		d_2 & 0 & 1 & 0 & 1  & 0& 0& 1& 0& 0& 0\\
		d_3 & 1 & 0 & 1 & 0  & 0& 0& 0& 1& 1& 0\\ 	
	}
	\]

In DTM, each document has a limited number of words and doesn't cover all the words in a corpus. So, most elements are zero in a row.  DTM suffers from two problems: sparsity and high dimensionality \citep{peng2018subspace}. Sparsity means that the number of elements having zero value is more than the number of elements having non-zero value \citep{karami2017taming}. High dimensionality denotes that there are numerous elements in DTM leading to cost and time concerns \citep{aggarwal2012introduction}. For instance, a DTM with 50,000 documents and 100,000 words has 5 billion elements where more than 80\% of them is zero. These two problems have negative effects on the performance of data mining techniques \citep{crain2012dimensionality}. 

To overcome sparsity and high dimensionality problems, dimension reduction (DR) methods have been developed as a per-processing step for reducing the original DTM dimensionality, and improving speed and accuracy of machine learning methods. Four DR strategies have been developed: supervised-feature selection (SFS), unsupervised-feature selection (UFS), supervised-feature transformation (SFT), and unsupervised-feature transformation (UFT). Although the supervised approach uses class labels for the learning task, there is no prior knowledge for the unsupervised approach \citep{liu2007computational}. Feature selection finds a subset of words in DTM that can describe the original data for supervised or unsupervised learning tasks \citep{wu2002feature,ABUALIGAH201724}. Many existing databases are unlabeled because large amounts of data make it difficult for humans to manually label the categories of each document. Moreover, manual labelling is expensive and subjective. Hence, unsupervised learning is needed for DR to minimize expense and time for the learning task runtime. It is worth mentioning that unsupervised DR methods can also be used for supervised learning tasks. On the other side, feature transformation methods use all the original data points, but feature selection methods use only a subset of the original data points. Because a large portion of data is lost by feature selection methods, UFT-based DR methods are preferred \citep{mac2013unsupervised}.

DR methods using the UFT strategy are based on approaches such as linear algebra, statistical distributions, and neural networks \citep{van2009dimensionality}. Although the fuzzy approach has contributed to DR based on the feature selection approach  \citep{jensen2004semantics,mac2013unsupervised}, fuzzy clustering has not been considered as a DR approach. 

This paper will investigate the DR ability of fuzzy clustering along with the performance impact of using different global term weighting (GTW) methods and fuzzifier values. This research compares the DR application of fuzzy clustering with principal component analysis (PCA) and singular value decomposition (SVD) using document classification and shows that fuzzy clustering has computational advantages over PCA and SVD. Specifically, the goals of this research with respect to DR application are to:

\begin{itemize}
	\item compare the performance of fuzzy clustering, PCA, and SVD with and without applying GTW methods.
	\item explore the impact of different fuzzifier values on fuzzy clustering performance.  
\end{itemize}

The contributions of this manuscript are three-fold. First, this paper introduces fuzzy clustering as a new approach for text data dimensionality reduction. Second, this study explores different fuzzifier values to propose a proper fuzzifier value for the dimensionality reduction application. Third, this research investigates the impact of GTW methods on fuzzy clustering. 

The remainder of this paper is organized as follows. In the related work section, we review the DR research. In the methodology and experiment sections, we provide more details about using fuzzy clustering as a new DR approach along with an evaluation study to verify the effectiveness of fuzzy clustering. Finally, we present a summary, limitations, and future directions in the last section.

\section{Related Work}

\begin{figure*}[htp!]
	\centering
	\large
	\scalebox{0.58}{
		
		\begin{tikzpicture}[grow=down, -stealth]
		
		\node[bag]{Dimension Reduction} 
		child{ edge from parent node[left]{}; \node[bag]{Supervised}
			child[missing]
			child{ edge from parent node[left]{}; \node[bag]{Feature Selection}}
			child[missing]
			child{ edge from parent node[right]{}; \node[bag]{Feature Transformation}}
			child[missing]
		}
		child{ edge from parent node[right]{}; \node[bag]{Unsupervised}
			child[missing]
			child{ edge from parent node[left]{}; \node[bag]{Feature Selection}}
			child[missing]
			child{ edge from parent node[right]{}; \node[bag]{Feature Transformation}}
			child[missing]
		};

		\end{tikzpicture}}
	\caption{Dimension Reduction Strategies}
	\label{fig:drstrategies}
\end{figure*}
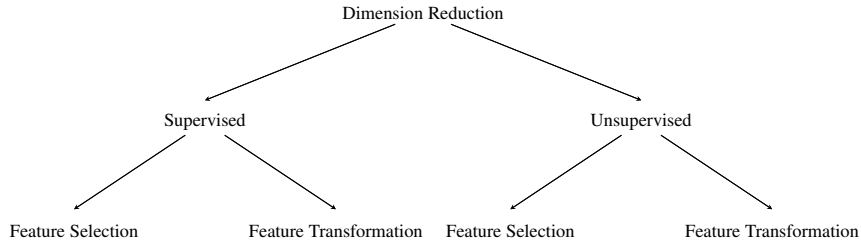

Using large volumes of text data has encouraged researchers to look for DR methods that enhance the quality of the original documents with a lower dimensional representation \citep{crain2012dimensionality,saarikoski2015influence}. Many DR methods have been applied to text mining based on four categories \citep{cunningham2008dimension}: supervised-feature selection (SFS), unsupervised-feature selection (UFS), supervised-feature transformation (SFT), and unsupervised-feature transformation (UFT) (Fig. \ref{fig:drstrategies}). Feature transformation methods use all the original words, but feature selection methods use only a subset of the original words. Unlike supervised learning, which uses class labels to help DR, it is a difficult problem in unsupervised learning to reduce the original dimensionality without the class labels \citep{dy2004feature}. In this section, we review these four DR strategies.

\subsection{Supervised-Feature Selection}
SFS explores the best minimum subset of the original words (features) for labeled data. Assume that $W=\{w_1,w_2,...,w_m\}$ and $L=\{l_1,l_2,...,l_k\}$ denote the words and the class label set where $m$ and $k$ are the number of words and labels, respectively. $D=\{d_1,d_2,...,d_n\}$ is the corpus where $n$ is the number of documents. The goal of supervised-feature selection strategy is to find $F=\{f_1,f_2,...,f_p\}$ that is a subset of $W$ with $p$ features ($p < m$) with respect to $L$. The most popular SFS-based methods are information gain \citep{yang1997comparative}, ReliefF \citep{liu2007computational}, Chi-square measure \citep{gao2017learning}, and genetic algorithm \citep{goldberg1988genetic}.

\subsection{Supervised-Feature Transformation}

SFT categories transfer the words to new dimensions for labeled documents. The goal of SFT strategy is to map the words in $W$ onto $q$ clusters, $C=\{c_1,c_2,...,c_q\}$, with respect to $L$ where $q << m$. For example, linear discriminant analysis is a supervised-feature transformation method using Fisher criterion based on maximizing the between class scatter and minimizing the within class scatter \citep{mika1999fisher}. 

\subsection{Unsupervised-Feature Selection}

Unlike SFS, UFS explores the best minimum subset of the original words for unlabeled data. The goal of unsupervised-feature selection strategy is to find $F$ that is a subset of $W$ with $p$ features ($p < m$) without having $L$.  Different methods have developed based on UFS strategy such as non-negative matrix factorization \citep{lee1999learning}, Laplacian score \citep{he2006laplacian},  category utility \citep{GluckCorter1985InfoUncertUtil}, and expectation maximisation \citep{dy2004feature}.        

\subsection{Unsupervised-Feature Transformation}

Unlike SFT, UFT transfers the words to new dimensions for unlabeled data. The goal of UFT strategy is to map the words in $W$ onto $q$ clusters, $C=\{c_1,c_2,...,c_q\}$, without having $L$ where $q << m$. For instance,  principal components analysis (PCA) is a linear unsupervised-feature transformation to map a set of correlated features into a set of uncorrelated features using orthogonally \citep{abdi2010principal,rajput2012iqram}. Although other UFT-based methods such as Kernel PCA \citep{scholkopf1998nonlinear}, Isomap \citep{tenenbaum2000global}, Laplacian Eigenmaps \citep{belkin2002laplacian}, local tangent space analysis \citep{zhang2004principal}, and multilayer autoencoders \citep{hinton2006reducing}, PCA is still among the most effective UFT-based method \citep{van2009dimensionality}. Latent semantic analysis (LSA) is another popular UFT-based method. While PCA uses eigen-decomposition of the covariance matrix, LSA uses singular value decomposition (SVD) for feature transformation \citep{deerwester1990indexing}. SVD detects the maximum variance of the data in a set of orthogonal basis vectors \citep{sweeney2014comparison}.

\textit{Feature selection vs feature transformation methods.} The output of feature selection methods is easy to interpret; however, the output of feature transformation may not be interpreted by the domain expert \citep{cunningham2008dimension}. While the feature transofmration methods use all the features of a datasets, the information loss is a side effect of the feature selection methods during the feature picking process \citep{ververidis2009information}.

\textit{Supervised vs unsupervised methods.} While the objective of supervised mtehods is clear, the objective of unsupervised methods is less clear \citep{cunningham2008dimension}. In addition, the number of categories is known for supervised methods; however, that number is unknown for unsupervised methods. Due to the labeling process, the supervised process is costly and time-consuming; however, the unsupervised process is inexpensive and efficient \citep{hindawi2013feature}. 

Some studies have used fuzzy logic to develop DR methods using supervised and unsupervised feature selection strategies such as the methods developed based on rough set attribute reduction (RSAR) \citep{jensen2004semantics,jensen2004fuzzy,chouchoulas2001rough,jelonek1995rough,shen2001rough}. These methods rely on retaining important features and removing irrelevant and redundant (noisy) features \citep{mac2013unsupervised}. This research investigates the application of fuzzy clustering for DR based on a UFT strategy to avoid losing information.

\section{Method}

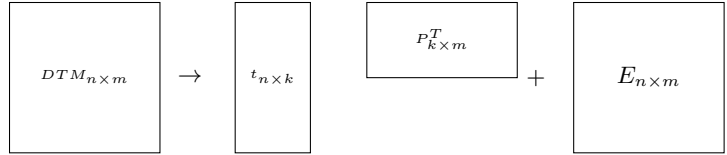
\begin{figure}[ht]
	\centering
	\begin{tikzpicture}
	\tiny

	\draw (0,-1) rectangle (2,1) node[pos=.5] {$DTM_{n \times m}$};
	\small \draw (2.4,0) node {$\rightarrow$};
	\tiny
	\draw (3,-1) rectangle (4,1) node[pos=.5] {$t_{n \times k}$};
	\draw (4.75,0) rectangle (6.75,1) node[pos=.5] {$P^T_{k \times m}$};
	
	\small \draw (7,0) node {$+$};
	
	\draw (7.5,-1) rectangle (9.5,1) node[pos=.5] {$E_{n \times m}$};
	
	
	\end{tikzpicture} 
	\caption{Matrix Interpretation of PCA \citep{bro2014principal}}
	\label{tab:pca}
\end{figure}

This paper uses fuzzy clustering for DR based on a UFT strategy to obtain a new basis that is an optimum combination of the original bases for unlabeled documents. Among the related well-known methods, PCA and LSA are widely used methods \citep{hinton2006reducing}. PCA converts $DTM$ that contains $n$ objects or documents with $m$ variables or words to three matrices: a linear combination of variables for each object ($t$), vectors of regression coefficients ($P$), and residuals ($E$) (Fig. \ref{tab:pca}).

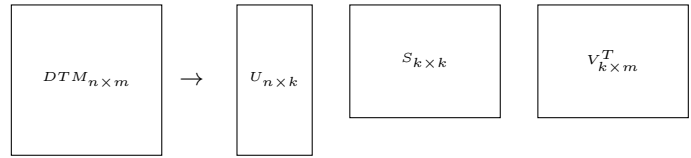
\begin{figure}[ht]
	\centering
	\begin{tikzpicture}
	\tiny

	\draw (0,-1) rectangle (2,1) node[pos=.5] {$DTM_{n \times m}$};
	\small \draw (2.4,0) node {$\rightarrow$};
	\tiny
	\draw (3,-1) rectangle (4,1) node[pos=.5] {$U_{n \times k}$};
	
	\draw (4.5,-0.5) rectangle (6.5,1) node[pos=.5] {$S_{k \times k}$};

	\draw (7,-0.5) rectangle (9,1) node[pos=.5] {$V^T_{k \times m}$};
	
	
	\end{tikzpicture} 
	\caption{Matrix Interpretation of SVD \citep{cunningham2008dimension}}
	\label{tab:svd}
\end{figure}

On the other hand, LSA applies SVD to $DTM$ to drop the least significant singular values and to keep $k$ singular values. SVD converts $DTM$ to three matrices: diagonalise $DTM \times DTM^T (U)$, singular values of $DTM (S)$, and diagonalise $DTM^T \times DTM (V^T)$ (Fig. \ref{tab:svd}). In both PCA and SVD, the original basis is represented by a new reduced base with k dimensions where $d<<m$ and $d<<n$.

The traditional reasoning has a precise character that uses true-or-false rather than more-or-less decisions \citep{zimmermann2010fuzzy}. Fuzzy logic adds a new extension to this reasoning moving from the classical logic of 0 or 1 to the truth values \textit{between} zero and one \citep{zadeh1973outline}. In fuzzy logic, if $X$ is a collection of data points represented by $x$, then a fuzzy set $A$ in $X$ is a set of ordered pairs, $ A=\{(x,\mu_A (x)|x \in X)\}$. $\mu_A(x)$ is the membership function which maps $X$ to the membership space $M$, which is between 0 and 1 \citep{karami2012fuzzy}. 

Clustering is an unsupervised approach for grouping similar documents \citep{siddiky2012data,jayabharathy2015correlation}. The goal of most clustering algorithms is to minimize the objective function ($J$) that measures the quality of clusters to find the optimum $J$ which is the sum of the squared distances between each cluster center and each data point \citep{ahmed2018clustering}. There are two major clustering approaches: hard and fuzzy (soft). The hard approach assigns exactly one cluster to a document, but the fuzzy approach assigns a degree of membership with respect to each of cluster for a document \citep{karami2015fuzzy}. Among fuzzy clustering techniques, fuzzy C-means (FCM) is the most popular model \citep{bezdek1981pattern} that minimizes $J$ by considering the following constraints: 

\begin{equation}
Min \: \: J_q =\sum_{f=1}^{k} \sum_{j=1}^{n} (\mu_{fj})^q ||d_j-v_f||^2         
\end{equation}
subject to:
\begin{equation}
0 \leq \mu_{fj}\leq1;
\end{equation}
\begin{equation}
\sum_{f=1}^{c} \mu_{fj}=1
\end{equation}
\begin{equation}
0<\sum_{j=1}^{n} \mu_{fj} < n; 
\end{equation}
Where: \\

\noindent $n$= number of documents\\
$k$= number of clusters\\
$\mu$= membership value\\
$q$= fuzzifier, $1 < q \le \infty$ \\
$d$= document vector\\
$v$= cluster center vector\\

\begin{figure}[ht!]
	\centering
	\begin{tikzpicture}
	\small
	
	\draw (-2.2,0) node [rotate=90] {Documents};
	\draw (0,1.2) node {Words};
	
	\draw (4,1.2) node {Fuzzy Clusters};
	
	\draw (2.75,0) node [rotate=90] {Documents};

	\draw (-2,-1) rectangle (2,1) node[pos=.5] {$DTM_{n \times m}$};
	\small \draw (2.4,0) node {$\rightarrow$};
	\tiny
	\draw (3,-1) rectangle (5,1) node[pos=.5] {$C_{n \times k}$};

	
	\end{tikzpicture}

	\caption{Matrix Interpretation of FC}
	\label{tab:FC}
\end{figure}
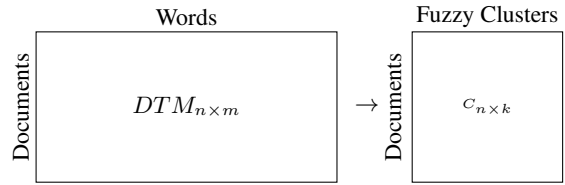

In this research, we use fuzzy clustering to cluster the documents represented by $DTM$ in a fuzzy way. The membership degree for each document with respect to each of the clusters is between 0 and 1 and is assumed to be a new basis to represent $DTM$. We assume that fuzzy clustering converts $DTM$ with $n$ documents and $m$ words to a new reduced matrix ($C$) with $k$ variables or dimensions ($k<<m$) (Fig. \ref{tab:FC}). Also, fuzzy clustering does not lose information in $DTM$ and does not need to select a subset of dimensions such as in SVD.

Assume that there are 10 words in a corpus with 5 documents represented by a $DTM$ (Fig. \ref{fig:fcexm}). Word 4 ($w_4$), for instance, appears two times in document 3 ($d_3$).   By applying fuzzy clustering on $DTM$ to find two fuzzy clusters,  matrix $C$ is created. For instance, document 5 ($d_5$) with a membership value of $0.4106981$  belongs to cluster 1 ($c_1$) and with a membership value of $0.5893019$ belongs to cluster 2. In this example, fuzzy clustering converts $DTM_{5 \times 10}$ sparse matrix to a non-sparse $C_{5 \times 2}$  matrix and reduces the dimension space by 80\%. 

\begin{figure*}[htp!]
	\[
	DTM= \kbordermatrix{
		& w_1 & w_2 & w_3 & w_4 & w_5 & w_6 & w_7 & w_8 & w_9 & w_{10} \\
		d_1 & 1& 0& 0& 1& 0& 0& 1& 2& 1&0\\
		d_2 & 2& 0& 1& 0& 0& 1& 0& 0& 0& 1 \\
		d_3 & 1& 0& 0& 2& 1& 0& 0& 1& 1&0 \\
		d_4 & 1& 1& 0& 0& 0& 1& 1& 1& 0&1 \\
		d_5 & 0& 0& 0& 1& 0& 1& 0& 0& 0& 0 
	} \rightarrow C=\kbordermatrix{
	& c_1 & c_2  \\
	d_1 & 0.2118281 & 0.7881719 \\ 
	d_2 & 0.8619096 & 0.1380904 \\
	d_3 & 0.0681949 & 0.9318051 \\
	d_4 & 0.8301873 & 0.1698127 \\
	d_5 & 0.4106981 & 0.5893019	
}
\]

\caption{A Numerical Example for DR Application of Fuzzy Clustering}
\label{fig:fcexm}
\end{figure*}

Numerous fuzzy clustering variations have been developed \citep{baraldi1999surveyI,baraldi1999surveyII}. In this study, we use the soft (fuzzy) spherical k-means method because this method has been developed for sparse text and can be efficiently parallelized for large scale datasets \citep{dhillon2001concept}. This method iterates between determining optimal memberships for fixed prototypes and computing optimal prototypes for fixed memberships \citep{dhillon2001concept}. 

This paper examines the DR performance of PCA, SVD, and fuzzy clustering (FC) with and without applying four GTW methods including \textit{entropy, inverse document frequency (IDF), probabilistic inverse document frequency (ProbIDF),} and \textit{normal} (Table~\ref{tab:gtw}).

\begin{table}[ht]
	\centering
	\caption{GTW Methods}
	{\begin{tabular}{c|c} 
			\textbf{Name} & \textbf{Formula} \\ \hline
			Entropy & $1+ \frac{\sum_j p_{ij}\log_2 (p_{ij})}{\log_2 n}$ \\
			& \\ 
			
			GFIDF  & $\frac{\sum_j f_{ij}}{\sum_j b(f_{ij})}$   \\
			& \\ 
			
			IDF & $\log_2 \frac{n}{\sum_j tf_{ij}}$   \\
			& \\ 
			
			Normal & $\frac{1}{\sqrt{\sum_j tf_{ij}^2}}$ \\

		\end{tabular}
		
		\label{tab:gtw}
	}
\end{table}

Symbol $tf_{ij}$ defines the number of times word $i$ occurs in document $j$. With $m$ words and $n$ documents:

\begin{equation}
b(tf_{ij})=\left\{
\begin{array}{c l}     
1 & tf_{ij}>0\\
0 & tf_{ij}=0
\end{array}\right.
\end{equation}

\begin{equation}
p_{ij}=\frac{tf_{ij}}{\sum_j tf_{ij}}
\end{equation}	

The entropy method gives higher weight to the terms that occur less frequently and in few documents \citep{dumais1992enhancing,karami2014fftm}. While IDF assigns higher weights to rare terms and lower weights to common terms \citep{papineni2001inverse,patrick2015textual}, GFIDF is another IDF variation in which words appearing once in every document or once in one document get the smallest weight \citep{karami2015fuzzy,karami2015fuzzyiconf}. ProbIDF gives weight to words based on frequency in one document and in all documents \citep{dumais1992enhancing}. Finally, the normal method is used to correct discrepancies in document lengths and to normalize the document vectors \citep{kolda1998limited,karami2015flatm}.

\section{Experiments}

This section is divided into two parts that consider the results generated by fuzzy clustering, PCA, and SVD for DR purpose with and without applying GTW methods. The goal of the experiments is to evaluate the DR application of fuzzy clustering against PCA and SVD by document classification. This research considers the accuracy average of three high performance classification algorithms including Adaptive Boosting (AdaBoost), Functional Trees (FT), and Random Forest \citep{chimieski2013association,sweeney2014comparison,rao2015performance,wu2008top,caruana2006empirical,qi2006evaluation}. This evaluation and complexity analysis help to demonstrate the usefulness of the DR application of fuzzy clustering for high dimensional sparse text data. 

This study uses two datasets, the \textit{irbla} R package for computing SVD and PCA \citep{Baglama2017irlba}, the skmeans R package for fuzzy spherical k-means clustering with 100 iterations and 1e-5 as the minimum improvement in objective function between two consecutive iterations \citep{Hornik2017skmeans}, the \textit{lsa} R package for computing GTW methods \citep{Wild2015lsa}, and the \textit{Weka} tool \cite{hall2009weka} with its default settings for document classification. The experiments consist of three steps: data preparation, dimension reduction, and classifier learning. The two benchmark datasets with thousands of documents employed between 10 and 100 reduced (transferred) features (dimensions).

\subsection{Datasets}

In this research, we leverage the two publicly available datasets:

\begin{itemize}
	\item Reuters dataset \citep{reutersdataset}: This dataset has 21,578 documents identified  with different news categories. This corpus has been utilized in several studies such as \cite{karami2017taming}, \cite{wang2018coupled}, and \cite{revanasiddappa2019new}. Two classes were created for binary classification. The documents in the Grain class were labelled as ``Grain" and the rest of the documents were labeled as ``Not Grain".

	\item Ohsumed dataset \citep{hersh1994ohsumed}: This dataset has 20,000 documents with different cardiovascular disease categories. Two classes were created for binary classification. This corpus has been used in different studies such as \cite{karami2018fuzzy}, \cite{kim2018towards}, and \cite{kim2019multi}. The documents in the Virus Disease class were labeled as ``Virus Diseases" and  5000 documents were randomly selected from the rest of the documents and labeled as ``Not Virus Diseases". 
\end{itemize}

\subsection{Document Classification}

Document classification assigns a document to a class using the words as features in a corpus. To avoid high dimensionality and sparsity, fuzzy clustering, SVD, and PCA are used to reduce the number of features without considering the labels. We then trained the three classification methods on ten reduced dimensions from 10 to 100, incremented by 10. 

To avoid optimistically biased sampling with respect to the whole dataset, this research tracks the performance of fuzzy clustering against PCA and SVD with the 5-fold cross validation method where the data is broken into five subsets for five iterations. Then these five results are combined to create a single estimation. The benefit of this method is that all documents are used for both training and testing. Each of the subsets is selected for testing and the rest of the sets are selected for training. 

The output of a classifier is presented as a confusion matrix (Table \ref{tab:confmx}) with the following definitions \citep{tsai2011intelligent,rinaldi2013multimodal}:

\begin{table}[ht]
	\centering
	\caption{Confusion Matrix}
	{\begin{tabular}{cc|c|c|c|}
			\cline{3-4}
			& & \multicolumn{2}{ c| }{\textbf{Predicted}} \\ \cline{3-4}
			& & \textbf{\textit{Negative}} & \textbf{\textit{Positive}}  \\ \cline{1-4}
			\multicolumn{1}{ |c| }{\multirow{2}{*}{\textbf{Actual}} } &
			\multicolumn{1}{ |c| }{\textbf{\textit{Negative}}} & TN & FP    \\ \cline{2-4}
			\multicolumn{1}{ |c  }{}                        &
			\multicolumn{1}{ |c| }{\textbf{\textit{Positive}}} & FN & TP      \\ \cline{1-4}
			
		\end{tabular}

		\label{tab:confmx}}
\end{table}

\begin{itemize}
	\item True Negative (TN) is the number of correct predictions that an instance is negative.
	
	\item False Negative (FN) is the number of incorrect predictions that an instance negative.
	
	\item False Positive (FP) is the number of incorrect predictions that an instance is positive.
	
	\item True Positive (TP) is the number of correct predictions that an instance is positive.
\end{itemize}

This step comes with and without applying four GTW methods to determine whether these methods offer an improved classification performance over the data. Classification accuracy is an evaluation metric to measure how well the classifier recognizes instances of the various classes. The accuracy of a classifier is the percentage of correctly classified documents in a test set \citep{chimieski2013association}. 

\begin{equation} 
Accuracy = \frac{TP+TN}{TP+TN+FP+FN}
\end{equation}

\subsection{Evaluation Results}

\begin{figure*}[htp!]
	\centering

	\begin{tikzpicture}[scale=.75]
	\begin{axis}
	[xlabel=Number of Dimensions,ylabel= Accuracy,legend style={at={(0.5,-0.2)}, anchor=north,legend columns=-1},
	y tick label style={
		/pgf/number format/.cd,
		fixed,
		fixed zerofill,
		precision=5,
		/tikz/.cd
	}
	]
	
	\addplot coordinates { (10,0.95531) (20,0.95528) (30,0.95531) (40,0.95519)(50,0.95519)(60,0.95525)(70,0.95519)(80,0.95525)(90,0.95519)};

	\addplot coordinates { (10,0.95525)(20,0.95519)(30,0.95525) (40,0.95519)(50,0.95528)(60,0.95522) (70,0.95504) (80,0.95528)(90,0.95522)};
	
	\addplot coordinates { (10,0.95516)(20,0.95525) (30,0.95519)(40,0.95525)(50,0.95516)(60,0.95478) (70,0.95478) (80,0.95475) (90,0.95463)};
	
	\addplot coordinates { (10,0.95531) (20,0.95525)(30,0.95519) (40,0.95495)(50,0.95410)(60,0.95484) (70,0.95472)(80,0.95442)(90,0.95433)};
	
	\addplot coordinates { (10,0.95388) (20,0.95283) (30,0.95274)(40,0.95338)(50,0.95326)(60,0.95333)(70,0.95344)(80,0.95336) (90,0.95330)};
	
	\addplot coordinates { (10,0.95374)(20,0.95271)(30,0.95274)(40,0.95304)(50,0.95307)(60,0.95310) (70,0.95298) (80,0.95316) (90,0.95319)};
	
	\legend{$FC-1.5$,$FC-2$,$FC-2.5$,$FC-3$,$PCA$, $SVD$}
	
	\end{axis}
	\end{tikzpicture}

	\caption{Classification Evaluation for Reuters Dataset}
	\label{fig:RAcc}
\end{figure*}
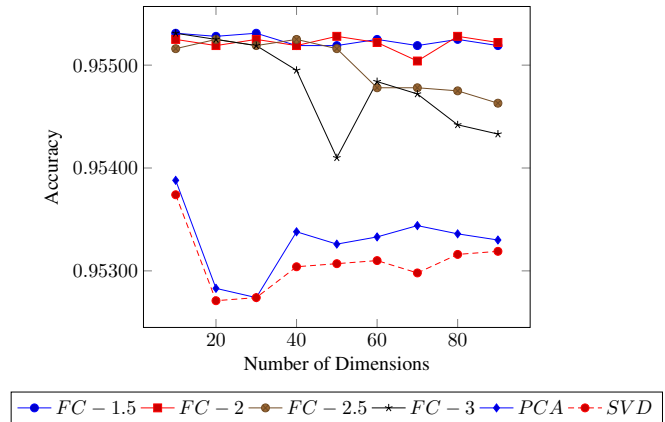

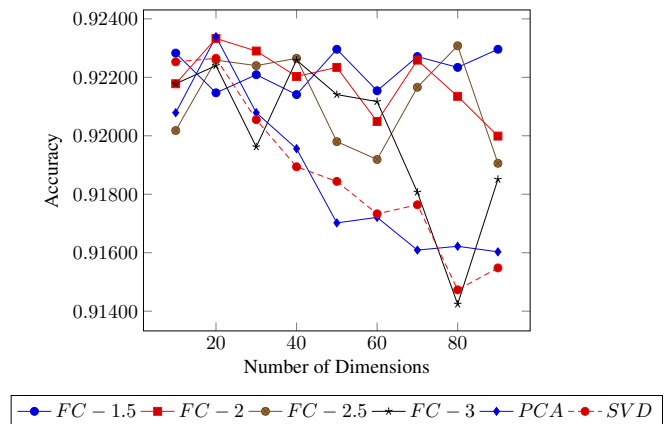
\begin{figure*}[htp!]
	\centering
	\begin{tikzpicture}[scale=.75]
	\begin{axis}
	[xlabel=Number of Dimensions,ylabel= Accuracy,legend style={at={(0.5,-0.2)}, anchor=north,legend columns=-1},
	y tick label style={
		/pgf/number format/.cd,
		fixed,
		fixed zerofill,
		precision=5,
		/tikz/.cd
	}
	]
	\addplot coordinates { (10,0.92283)(20,0.92147)(30,0.92209)(40,0.92141) (50,0.92296) (60,0.92154)(70,0.92271)(80,0.92234) (90,0.92296)};
	
	\addplot coordinates { (10,0.92178)(20,0.92333)(30,0.92290) (40,0.92203) (50,0.92234) (60,0.92049)(70,0.92259)(80,0.92135) (90,0.91999)};
	
	\addplot coordinates { (10,0.92018)(20,0.92259)(30,0.92240) (40,0.92265) (50,0.91980) (60,0.91919)(70,0.92166)(80,0.92308) (90,0.91906)};
	
	\addplot coordinates { (10,0.92178)(20,0.92240)(30,0.91962) (40,0.92259) (50,0.92141) (60,0.92117)(70,0.91807)(80,0.91424) (90,0.91851)};
	
	\addplot coordinates { (10,0.92079) (20,0.92339)(30,0.92079)(40,0.91956) (50,0.91702) (60,0.91721)(70,0.91609)(80,0.91622) (90,0.91603)};
	
	\addplot coordinates { (10,0.92253)(20,0.92265)(30,0.92055)(40,0.91894) (50,0.91844) (60,0.91733)(70,0.91764)(80,0.91473) (90,0.91548)};
	
	\legend{$FC-1.5$,$FC-2$,$FC-2.5$,$FC-3$,$PCA$, $SVD$}
	\end{axis}
	
	\end{tikzpicture}

	\caption{Classification Evaluation for OHSUMED Dataset}
	\label{fig:OhAcc}
\end{figure*}

In this part, we report the evaluation results based on the document classification of the two datasets. Fig. \ref{fig:RAcc} and Fig. \ref{fig:OhAcc} show the average of the three classifiers accuracy for PCA, SVD, and fuzzy clustering with four fuzzifier values including $q$=1.5 ($FC-1.5$), $q$=2 ($FC-2$), $q$=2.5 ($FC-2.5$), and $q$=3 ($FC-3$). From these two figures, it can be seen that DR using fuzzy clustering produces better results than DR using PCA and SVD for all the number of dimensions from 10 to 100.

FC-1.5 provides better accuracy in most of the classification experiments and shows the highest stability with the lowest standard deviation value followed by FC-2, FC-2.5, FC-3, SVD, and PCA. Although increasing the number of dimensions mostly has a negative effect primarily on the accuracy performance of PCA and SVD, fuzzy clustering shows a stable performance with a lower standard deviation than SVD and PCA. While SVD shows more stability than PCA, the latter exhibits better accuracy than the earlier one. 

\begin{figure*}[htp!]
	\centering
	\begin{tikzpicture}[scale=0.8]
	\begin{axis}[
	ybar,
	width  = 16cm,
	height = 4cm,
	symbolic x coords = {FC-1.5, FC-2,FC-2.5,FC-3, PCA, SVD },
	xtick={FC-1.5, FC-2,FC-2.5,FC-3, PCA, SVD},
	axis y line*=left,
	axis x line*=bottom,
	legend style={at={(0.5,-0.25)},
		anchor=north,legend columns=-1},
	y tick label style={
		/pgf/number format/.cd,
		fixed,
		fixed zerofill,
		precision=5,
		/tikz/.cd
	}
	]
	\addplot[black,fill=yellow,pattern=north east lines] coordinates {(FC-1.5,0.95531) (FC-2,0.95525) (FC-2.5,0.95516) (FC-3,0.95531) (PCA,0.95388) (SVD,0.95374)};
	\addplot[black,fill=yellow,pattern=horizontal lines] coordinates {(FC-1.5,0.95531) (FC-2,0.95525) (FC-2.5,0.95528) (FC-3,0.95522) (PCA,0.95374) (SVD,0.95401)};
	\addplot[black,fill=yellow,pattern=vertical lines] coordinates {(FC-1.5,0.95522) (FC-2,0.95522) (FC-2.5,0.95525) (FC-3,0.95522) (PCA,0.95415) (SVD,0.95424)};
	\addplot[black,fill=yellow,pattern=north west lines] coordinates {(FC-1.5,0.95528) (FC-2,0.95528) (FC-2.5,0.95525) (FC-3,0.95522) (PCA,0.95380) (SVD,0.95399)};
	\addplot[black,fill=yellow,pattern=grid] coordinates {(FC-1.5,0.95525) (FC-2,0.95537) (FC-2.5,0.95522) (FC-3,0.95525) (PCA,0.95410) (SVD,0.95404)};
	
	\legend{No-GW,Entropy,GF-IDF,IDF,Normal}
	\end{axis}
	\end{tikzpicture}
	\caption{Classification Evaluation for Reuters Dataset with GW - K=10}
	\label{fig:k10}

	\begin{tikzpicture}[scale=0.8]
	\begin{axis}[
	ybar,
	width  = 16cm,
	height = 4cm,
	symbolic x coords = {FC-1.5, FC-2,FC-2.5,FC-3, PCA, SVD },
	xtick={FC-1.5, FC-2,FC-2.5,FC-3, PCA, SVD},
	axis y line*=left,
	axis x line*=bottom,
	legend style={at={(0.5,-0.25)},
		anchor=north,legend columns=-1},
	y tick label style={
		/pgf/number format/.cd,
		fixed,
		fixed zerofill,
		precision=5,
		/tikz/.cd
	}
	]
	
	\addplot[black,fill=yellow,pattern=north east lines] coordinates {(FC-1.5,0.955279889)  (FC-2,0.955191)  (FC-2.5,0.955250111)  (FC-3,0.955250111)  (PCA,0.952829444)  (SVD,0.952710889) };
	\addplot[black,fill=yellow,pattern=horizontal lines] coordinates {(FC-1.5,0.955043)  (FC-2,0.955161667)  (FC-2.5,0.955339)  (FC-3,0.955161667)  (PCA,0.953540111)  (SVD,0.953295778) };
	\addplot[black,fill=yellow,pattern=vertical lines] coordinates {(FC-1.5,0.955309667)  (FC-2,0.955220778)  (FC-2.5,0.955191)  (FC-3,0.955309667)  (PCA,0.953503222)  (SVD,0.953303333) };
	\addplot[black,fill=yellow,pattern=north west lines] coordinates {(FC-1.5,0.955161667)  (FC-2,0.955191)  (FC-2.5,0.955339)  (FC-3,0.955279889)  (PCA,0.952466222)  (SVD,0.953103) };
	\addplot[black,fill=yellow,pattern=grid] coordinates {(FC-1.5,0.955161667)  (FC-2,0.955250111)  (FC-2.5,0.955398556)  (FC-3,0.955043)  (PCA,0.952718)  (SVD,0.953562111) };

	\legend{No-GW,Entropy,GF-IDF,IDF,Normal}
	\end{axis}
	\end{tikzpicture}
	\caption{Classification Evaluation for Reuters Dataset with GW - K=20}
	\label{fig:k20}
	
	\begin{tikzpicture}[scale=0.8]
	\begin{axis}[
	ybar,
	width  = 16cm,
	height = 4cm,
	symbolic x coords = {FC-1.5, FC-2,FC-2.5,FC-3, PCA, SVD },
	xtick={FC-1.5, FC-2,FC-2.5,FC-3, PCA, SVD},
	axis y line*=left,
	axis x line*=bottom,
	legend style={at={(0.5,-0.25)},
		anchor=north,legend columns=-1},
	y tick label style={
		/pgf/number format/.cd,
		fixed,
		fixed zerofill,
		precision=5,
		/tikz/.cd
	}
	]

	\addplot[black,fill=yellow,pattern=north east lines] coordinates {(FC-1.5,0.955309667)  (FC-2,0.955250111)  (FC-2.5,0.955191)  (FC-3,0.955191)  (PCA,0.952740444)  (SVD,0.952740222) };
	\addplot[black,fill=yellow,pattern=horizontal lines] coordinates {(FC-1.5,0.955279889)  (FC-2,0.955220778)  (FC-2.5,0.955250111)  (FC-3,0.955279889)  (PCA,0.952910889)  (SVD,0.953207) };
	\addplot[black,fill=yellow,pattern=vertical lines] coordinates {(FC-1.5,0.955279889)  (FC-2,0.955220778)  (FC-2.5,0.955339)  (FC-3,0.955368778)  (PCA,0.953265889)  (SVD,0.953273556) };
	\addplot[black,fill=yellow,pattern=north west lines] coordinates {(FC-1.5,0.955279889)  (FC-2,0.955250111)  (FC-2.5,0.955279889)  (FC-3,0.955250111)  (PCA,0.952377778)  (SVD,0.953103) };
	\addplot[black,fill=yellow,pattern=grid] coordinates {(FC-1.5,0.955220778)  (FC-2,0.955250111)  (FC-2.5,0.955398556)  (FC-3,0.955309667)  (PCA,0.953162444)  (SVD,0.953651) };

	\legend{No-GW,Entropy,GF-IDF,IDF,Normal}
	\end{axis}
	\end{tikzpicture}
	\caption{Classification Evaluation for Reuters Dataset with GW - K=30}
	\label{fig:k30}
	
	\begin{tikzpicture}[scale=0.8]
	\begin{axis}[
	ybar,
	width  = 16cm,
	height = 4cm,
	symbolic x coords = {FC-1.5, FC-2,FC-2.5,FC-3, PCA, SVD },
	xtick={FC-1.5, FC-2,FC-2.5,FC-3, PCA, SVD},
	axis y line*=left,
	axis x line*=bottom,
	legend style={at={(0.5,-0.25)},
		anchor=north,legend columns=-1},
	y tick label style={
		/pgf/number format/.cd,
		fixed,
		fixed zerofill,
		precision=5,
		/tikz/.cd
	}
	]
	
	\addplot[black,fill=yellow,pattern=north east lines] coordinates {(FC-1.5,0.955191)  (FC-2,0.955191)  (FC-2.5,0.955250111)  (FC-3,0.954954111)  (PCA,0.953377222)  (SVD,0.953043889) };
	\addplot[black,fill=yellow,pattern=horizontal lines] coordinates {(FC-1.5,0.955220778)  (FC-2,0.955220778)  (FC-2.5,0.955339)  (FC-3,0.954539444)  (PCA,0.953066222)  (SVD,0.953347444) };
	\addplot[black,fill=yellow,pattern=vertical lines] coordinates {(FC-1.5,0.955220778)  (FC-2,0.955279889)  (FC-2.5,0.955309667)  (FC-3,0.955043)  (PCA,0.953562333)  (SVD,0.953154889) };
	\addplot[black,fill=yellow,pattern=north west lines] coordinates {(FC-1.5,0.955161667)  (FC-2,0.955309667)  (FC-2.5,0.955427889)  (FC-3,0.955013222)  (PCA,0.952910667)  (SVD,0.953221667) };
	\addplot[black,fill=yellow,pattern=grid] coordinates {(FC-1.5,0.955131889)  (FC-2,0.955398556)  (FC-2.5,0.955161667)  (FC-3,0.954895)  (PCA,0.953014444)  (SVD,0.953591889) };

	\legend{No-GW,Entropy,GF-IDF,IDF,Normal}
	\end{axis}
	\end{tikzpicture}
	\caption{Classification Evaluation for Reuters Dataset with GW - K=40}
	\label{fig:k40}
	
	\begin{tikzpicture}[scale=0.8]
	\begin{axis}[
	ybar,
	width  = 16cm,
	height = 4cm,
	symbolic x coords = {FC-1.5, FC-2,FC-2.5,FC-3, PCA, SVD },
	xtick={FC-1.5, FC-2,FC-2.5,FC-3, PCA, SVD},
	axis y line*=left,
	axis x line*=bottom,
	legend style={at={(0.5,-0.25)},
		anchor=north,legend columns=-1},
	y tick label style={
		/pgf/number format/.cd,
		fixed,
		fixed zerofill,
		precision=5,
		/tikz/.cd
	}
	]
	
	\addplot[black,fill=yellow,pattern=north east lines] coordinates {(FC-1.5,0.955191)  (FC-2,0.955279889)  (FC-2.5,0.955161667)  (FC-3,0.954095444)  (PCA,0.953258556)  (SVD,0.953073667) };
	\addplot[black,fill=yellow,pattern=horizontal lines] coordinates {(FC-1.5,0.955131889)  (FC-2,0.955309667)  (FC-2.5,0.954510111)  (FC-3,0.954835889)  (PCA,0.953036444)  (SVD,0.953791889) };
	\addplot[black,fill=yellow,pattern=vertical lines] coordinates {(FC-1.5,0.954628333)  (FC-2,0.955339)  (FC-2.5,0.955250111)  (FC-3,0.955043)  (PCA,0.953621444)  (SVD,0.953428778) };
	\addplot[black,fill=yellow,pattern=north west lines] coordinates {(FC-1.5,0.955161667)  (FC-2,0.955339)  (FC-2.5,0.954895)  (FC-3,0.954895)  (PCA,0.952584889)  (SVD,0.953310556) };
	\addplot[black,fill=yellow,pattern=grid]  coordinates {(FC-1.5,0.955131889)  (FC-2,0.955309667)  (FC-2.5,0.955013222)  (FC-3,0.955072778)  (PCA,0.952999667)  (SVD,0.953799) };

	\legend{No-GW,Entropy,GF-IDF,IDF,Normal}
	\end{axis}
	\end{tikzpicture}
	\caption{Classification Evaluation for Reuters Dataset with GW - K=50}
	\label{fig:k50}

	\label{fig:RGWAcc21}
\end{figure*}

\begin{figure*}[htp!]
	\centering
	\begin{tikzpicture}[scale=0.8]
	\begin{axis}[
	ybar,
	width  = 16cm,
	height = 4cm,
	symbolic x coords = {FC-1.5, FC-2,FC-2.5,FC-3, PCA, SVD },
	xtick={FC-1.5, FC-2,FC-2.5,FC-3, PCA, SVD},
	axis y line*=left,
	axis x line*=bottom,
	legend style={at={(0.5,-0.25)},
		anchor=north,legend columns=-1},
	y tick label style={
		/pgf/number format/.cd,
		fixed,
		fixed zerofill,
		precision=5,
		/tikz/.cd
	}
	]
	
	\addplot[black,fill=yellow,pattern=north east lines] coordinates {(FC-1.5,0.95525)  (FC-2,0.95522)  (FC-2.5,0.95478)  (FC-3,0.95484)  (PCA,0.95333)  (SVD,0.95310) };
	\addplot[black,fill=yellow,pattern=horizontal lines] coordinates {(FC-1.5,0.95466)  (FC-2,0.95522)  (FC-2.5,0.95487)  (FC-3,0.95537)  (PCA,0.95300)  (SVD,0.95382) };
	\addplot[black,fill=yellow,pattern=vertical lines] coordinates {(FC-1.5,0.95522)  (FC-2,0.95531)  (FC-2.5,0.95531)  (FC-3,0.95448)  (PCA,0.95362)  (SVD,0.95322) };
	\addplot[black,fill=yellow,pattern=north west lines] coordinates {(FC-1.5,0.95519)  (FC-2,0.95534)  (FC-2.5,0.95487)  (FC-3,0.95469)  (PCA,0.95261)  (SVD,0.95325) };
	\addplot[black,fill=yellow,pattern=grid] coordinates {(FC-1.5,0.95504)  (FC-2,0.95528)  (FC-2.5,0.95528)  (FC-3,0.95490)  (PCA,0.95300)  (SVD,0.95368) };

	\legend{No-GW,Entropy,GF-IDF,IDF,Normal}
	\end{axis}
	\end{tikzpicture}
	\caption{Classification Evaluation for Reuters Dataset with GW - K=60}
	\label{fig:k60}

	\begin{tikzpicture}[scale=0.8]
	\begin{axis}[
	ybar,
	width  = 16cm,
	height = 4cm,
	symbolic x coords = {FC-1.5, FC-2,FC-2.5,FC-3, PCA, SVD },
	xtick={FC-1.5, FC-2,FC-2.5,FC-3, PCA, SVD},
	axis y line*=left,
	axis x line*=bottom,
	legend style={at={(0.5,-0.25)},
		anchor=north,legend columns=-1},
	y tick label style={
		/pgf/number format/.cd,
		fixed,
		fixed zerofill,
		precision=5,
		/tikz/.cd
	}
	]

	\addplot[black,fill=yellow,pattern=north east lines] coordinates {(FC-1.5,0.95519)  (FC-2,0.95504)  (FC-2.5,0.95478)  (FC-3,0.95472)  (PCA,0.95344)  (SVD,0.95298) };
	\addplot[black,fill=yellow,pattern=horizontal lines] coordinates {(FC-1.5,0.95516)  (FC-2,0.95537)  (FC-2.5,0.95504)  (FC-3,0.95436)  (PCA,0.95273)  (SVD,0.95379) };
	\addplot[black,fill=yellow,pattern=vertical lines] coordinates {(FC-1.5,0.95516)  (FC-2,0.95448)  (FC-2.5,0.95487)  (FC-3,0.95451)  (PCA,0.95377)  (SVD,0.95325) };
	\addplot[black,fill=yellow,pattern=north west lines] coordinates {(FC-1.5,0.95516)  (FC-2,0.95484)  (FC-2.5,0.95516)  (FC-3,0.95435)  (PCA,0.95301)  (SVD,0.95378) };
	\addplot[black,fill=yellow,pattern=grid] coordinates {(FC-1.5,0.95510)  (FC-2,0.95519)  (FC-2.5,0.95507)  (FC-3,0.95401)  (PCA,0.95330)  (SVD,0.95386) };

	\legend{No-GW,Entropy,GF-IDF,IDF,Normal}
	\end{axis}
	\end{tikzpicture}
	\caption{Classification Evaluation for Reuters Dataset with GW - K=70}
	\label{fig:k70}
	
	\begin{tikzpicture}[scale=0.8]
	\begin{axis}[
	ybar,
	width  = 16cm,
	height = 4cm,
	symbolic x coords = {FC-1.5, FC-2,FC-2.5,FC-3, PCA, SVD },
	xtick={FC-1.5, FC-2,FC-2.5,FC-3, PCA, SVD},
	axis y line*=left,
	axis x line*=bottom,
	legend style={at={(0.5,-0.25)},
		anchor=north,legend columns=-1},
	y tick label style={
		/pgf/number format/.cd,
		fixed,
		fixed zerofill,
		precision=5,
		/tikz/.cd
	}
	]
	
	\addplot[black,fill=yellow,pattern=north east lines] coordinates {(FC-1.5,0.95525)  (FC-2,0.95528)  (FC-2.5,0.95475)  (FC-3,0.95442)  (PCA,0.95336)  (SVD,0.95316) };
	\addplot[black,fill=yellow,pattern=horizontal lines] coordinates {(FC-1.5,0.95513)  (FC-2,0.95531)  (FC-2.5,0.95490)  (FC-3,0.95410)  (PCA,0.95324)  (SVD,0.95373) };
	\addplot[black,fill=yellow,pattern=vertical lines] coordinates {(FC-1.5,0.95495)  (FC-2,0.95519)  (FC-2.5,0.95457)  (FC-3,0.95454)  (PCA,0.95374)  (SVD,0.95322) };
	\addplot[black,fill=yellow,pattern=north west lines] coordinates {(FC-1.5,0.95519)  (FC-2,0.95531)  (FC-2.5,0.95478)  (FC-3,0.95397)  (PCA,0.95287)  (SVD,0.95334) };
	\addplot[black,fill=yellow,pattern=grid] coordinates {(FC-1.5,0.95531)  (FC-2,0.95525)  (FC-2.5,0.95433)  (FC-3,0.95412)  (PCA,0.95303)  (SVD,0.95392) };

	\legend{No-GW,Entropy,GF-IDF,IDF,Normal}
	\end{axis}
	\end{tikzpicture}
	\caption{Classification Evaluation for Reuters Dataset with GW - K=80}
	\label{fig:k80}
	
	\begin{tikzpicture}[scale=0.8]
	\begin{axis}[
	ybar,
	width  = 16cm,
	height = 4cm,
	symbolic x coords = {FC-1.5, FC-2,FC-2.5,FC-3, PCA, SVD },
	xtick={FC-1.5, FC-2,FC-2.5,FC-3, PCA, SVD},
	axis y line*=left,
	axis x line*=bottom,
	legend style={at={(0.5,-0.25)},
		anchor=north,legend columns=-1},
	y tick label style={
		/pgf/number format/.cd,
		fixed,
		fixed zerofill,
		precision=5,
		/tikz/.cd
	}
	]
	
	\addplot[black,fill=yellow,pattern=north east lines] coordinates {(FC-1.5,0.95519)  (FC-2,0.95522)  (FC-2.5,0.95463)  (FC-3,0.95433)  (PCA,0.95330)  (SVD,0.95319) };
	\addplot[black,fill=yellow,pattern=horizontal lines] coordinates {(FC-1.5,0.95469)  (FC-2,0.95418)  (FC-2.5,0.95439)  (FC-3,0.95433)  (PCA,0.95336)  (SVD,0.95400) };
	\addplot[black,fill=yellow,pattern=vertical lines] coordinates {(FC-1.5,0.95507)  (FC-2,0.95457)  (FC-2.5,0.95445)  (FC-3,0.95434)  (PCA,0.95350)  (SVD,0.95358) };
	\addplot[black,fill=yellow,pattern=north west lines] coordinates {(FC-1.5,0.95519)  (FC-2,0.95501)  (FC-2.5,0.95445)  (FC-3,0.95415)  (PCA,0.95337)  (SVD,0.95355) };
	\addplot[black,fill=yellow,pattern=grid] coordinates {(FC-1.5,0.95487)  (FC-2,0.95501)  (FC-2.5,0.95451)  (FC-3,0.95401)  (PCA,0.95341)  (SVD,0.95389) };

	\legend{No-GW,Entropy,GF-IDF,IDF,Normal}
	\end{axis}
	\end{tikzpicture}
	\caption{Classification Evaluation for Reuters Dataset with GW - K=90}
	\label{fig:k90}
	
	\begin{tikzpicture}[scale=0.8]
	\begin{axis}[
	ybar,
	width  = 16cm,
	height = 4cm,
	symbolic x coords = {FC-1.5, FC-2,FC-2.5,FC-3, PCA, SVD },
	xtick={FC-1.5, FC-2,FC-2.5,FC-3, PCA, SVD},
	axis y line*=left,
	axis x line*=bottom,
	legend style={at={(0.5,-0.25)},
		anchor=north,legend columns=-1},
	y tick label style={
		/pgf/number format/.cd,
		fixed,
		fixed zerofill,
		precision=5,
		/tikz/.cd
	}
	]

	\addplot[black,fill=yellow,pattern=north east lines] coordinates {(FC-1.5,0.95519)  (FC-2,0.95492)  (FC-2.5,0.95415)  (FC-3,0.95394)  (PCA,0.95365)  (SVD,0.95325) };
	\addplot[black,fill=yellow,pattern=horizontal lines] coordinates {(FC-1.5,0.95498)  (FC-2,0.95472)  (FC-2.5,0.95451)  (FC-3,0.95454)  (PCA,0.95327)  (SVD,0.95406) };
	\addplot[black,fill=yellow,pattern=vertical lines] coordinates {(FC-1.5,0.95466)  (FC-2,0.95472)  (FC-2.5,0.95457)  (FC-3,0.95382)  (PCA,0.95374)  (SVD,0.95361) };
	\addplot[black,fill=yellow,pattern=north west lines] coordinates {(FC-1.5,0.95525)  (FC-2,0.95481)  (FC-2.5,0.95472)  (FC-3,0.95421)  (PCA,0.95349)  (SVD,0.95352) };
	\addplot[black,fill=yellow,pattern=grid]  coordinates {(FC-1.5,0.95492)  (FC-2,0.95475)  (FC-2.5,0.95460)  (FC-3,0.95387)  (PCA,0.95288)  (SVD,0.95415) };

	\legend{No-GW,Entropy,GF-IDF,IDF,Normal}
	\end{axis}
	\end{tikzpicture}
	\caption{Classification Evaluation for Reuters Dataset with GW - K=100}
	\label{fig:k100}

	\label{fig:RGWAcc22}
\end{figure*}

Fig. \ref{fig:k10} to Fig. \ref{fig:k100} show the average of the accuracy for the three classifiers with and without applying GTW methods using the Reuters dataset. These figures indicate that 46\% of the experiments including SVD and 63\% of the experiments excluding SVD show negative or neutral effects of GTW methods on the classifiers. Applying GTW methods has a 72\% and 100\% positive effect on the performance of PCA and SVD, respectively. The highest effect of GTW methods is on SVD, followed by FC-2.5 with 72\% positive effect. The weighting methods with the most negative effect is exhibited by FC-1.5 with 87\% negative effect followed by PCA, FC-3, and FC-2 with 57\%, 55\%, and 52\% negative effects, respectively. In sum: 

\begin{itemize}
	\item DR with fuzzy clustering outperforms PCA and SVD in most of the experiments. 
	\item Fuzzy clustering with fuzzifier value 1.5 shows better DR performance than Fuzzy clustering with fuzzifier values 2, 2.5, and 3.
	\item GTW methods generally provide no effect or a negative effect on the performance of DR using fuzzy clustering. 
	\item PCA produces a better DR performance than SVD but GTW methods help SVD more than PCA to produce better DR.  
	
\end{itemize}

While the complexities for PCA and SVD methods are $O(mnlog(k))$ and $O(mnlog(k)+(m+n)k^2)$, respectively \citep{halko2009finding},  the complexity of the fuzzy spherical k-means method is  $O(n+k)$ \citep{dhillon2002iterative} where $m$ is the number of words, $n$ is the number of documents, and $k$ is the number of dimensions or clusters. Other than the complexity advantage, there are other benefits for the DR application of fuzzy clustering including no loss of dimensions, the ability to estimate the number of clusters or dimensions with the methods such as silhouette index \citep{campello2006fuzzy} and Xie-Beni index \citep{xie1991validity}, and working with both discrete and continuous data.

\section{Conclusion}

There are a large number of documents in online environments such as digital libraries and social media. The first step in analyzing these huge corpora is to represent the text data with DTM; however, this technique suffers from high dimensionality and sparsity problems. To overcome these two problems, DTM should be processed with DR methods to reduce the dimensionality for better accuracy. The exponential growth of text data indicates that DR still needs improvement and new perspectives. DR methods have been developed based on four strategies among which UFT is a popular and efficient strategy. While a wide range of UFT-based DR methods has been developed, fuzzy clustering has not been considered as a DR approach.

This study discusses the DR application of fuzzy clustering based on the UFT strategy. This paper applies fuzzy clustering to DTM to represent a matrix in which the elements are the fuzzy membership degree values of documents with respect to clusters. The efficiency and effectiveness of fuzzy clustering for DR are demonstrated through complexity and classification accuracy comparison with PCA and SVD using two well-known corpora.

This paper's results illustrate that fuzzy clustering is a competitor to powerful methods such as PCA and SVD in the context of DR for document collections. Indeed, the advantages of fuzzy clustering include less complexity, no loss of information, and the ability to work with both discrete and continuous data. Moreover, there are already developed methods to estimate the optimum number of dimensions (fuzzy clusters).

This paper has some implications. First, the proposed dimension reduction approach can be used for structured data, semi-structured, and unstructured data. Second, the proposed approach can be used as a pre-processing step for both supervised and unsupervised machine learning techniques. Third, the presented fuzzy approach improves the speed and accuracy of big data mining processes. Fourth, this research is beneficial for a wide range of applications such as information retrieval, pattern recognition, data visualization, and microarray data analysis in genetics.

This research has two limitations. The first one is that this paper has studied one fuzzy clustering method. The second one is that this research has utilized two dimension reduction methods for the evaluation. We will investigate other fuzzy clustering and dimension reduction methods in our future study.




\bibliographystyle{apalike}

\bibliography{refrence} 
\end{document}